\pdfoutput=1
\documentclass[11pt]{article}

\usepackage{acl}

\usepackage{times}
\usepackage{latexsym}
\usepackage{graphicx} % Required for inserting images
\usepackage{booktabs} % for pretty tables
\usepackage{lipsum} % for gibberish text
\usepackage{stfloats}
\usepackage{tabularx}
\usepackage{soul}
\usepackage{adjustbox}
\usepackage{appendix}
\usepackage{ctable}
\usepackage{amssymb}
\usepackage{mathtools, nccmath}
\usepackage{amsmath}
\usepackage{stmaryrd}  
\usepackage{caption}
\usepackage{subcaption}
\usepackage{multirow}
\usepackage{epstopdf}
\usepackage{paralist} %compactitem
\usepackage[utf8]{inputenc}
\usepackage{multirow}
\usepackage{xstring}
\usepackage[T1]{fontenc}
\usepackage[utf8]{inputenc}

\usepackage{microtype}

\usepackage{inconsolata}

\title{WorryWords: Norms of Anxiety Association for over 44k English Words}

\author{Saif M. Mohammad  \\
  National Research Council Canada \\
  {\tt saif.mohammad@nrc-cnrc.gc.ca} }
  
\begin{document}
\maketitle
\begin{abstract}
Anxiety, the anticipatory unease about a potential negative outcome, 
is a common and beneficial human emotion. 
However, there is still much that is not known, such as how anxiety relates to our body 
 and how it manifests in language. This is especially pertinent given the increasing impact of anxiety-related disorders.
% Yet, there is growing evidence that the rapid changes in our environment over the last few decades has led to an epidemic of anxiety-associated disorders.
% This has underscored a marked uptick of interest in anxiety (both normal and clinical) in psychological, social science, and medical research. Yet, there is little that we know about how anxiety manifests in language. 
In this work,
we introduce \textit{WorryWords}, the first large-scale repository of manually derived word--anxiety associations for over 44,450 English words. We show that the anxiety associations are highly reliable.
% split-half reliability of 0.82 Spearman and 0.89 Pearson correlation. 
We use WorryWords to study the relationship between anxiety and other emotion constructs, as well as the rate at which children acquire anxiety words with age. Finally, we show that using WorryWords alone, one can accurately track the change of anxiety in streams of text.
The lexicon enables a wide variety of anxiety-related research % and practical applications 
in psychology, NLP, public health, and social sciences.
WorryWords (and its translations to over 100 languages) is freely available.\\[3pt] 
\url{http://saifmohammad.com/worrywords.html}

%, and digital humanities.
\end{abstract}

% "Anxietarium"
% "MindSpike: The Anxiety Lexicon"
% "FearFont: An Atlas of Anxiety"
% "The Anxious Almanac"
% "The Anxiety Atlas: Mapping Emotions"
% "WorryWords: A Lexicon of Anxiety"
% "Panic Papers: Exploring Anxiety"
% "The Anxiety Archive: Words Unveiled"
% "The Nervous Narrative: A Lexicon"
% "Anxious Antonyms: A Lexical Journey"
% "The Anxiety A to Z Guide"
% "Nervous Notions: A Lexicon of Anxiety"
% "The Anxiety Glossary: From A to Z"
% "Restless Reflections: An Anxiety Lexicon"
% "The Anxiety Lexicography Project"
% "Anxiety ABCs: Words and Meanings"
% "Uneasy Etymology: Mapping Anxiety"
% "The Worrier's Wordbook"
% "Trepidation Terminology: An Anxiety Lexicon"
% "The Anxious Lexeme: Exploring Emotions"

\section{Introduction}

Anxiety is the apprehensive uneasiness or nervousness due to some anticipated or potential negative outcome. 
% It is a pervasive and central human emotion \cite{zeidner2010anxiety,salecl2004anxiety}. 
Even though it is often considered to be an unwanted and negative feeling, psychologists have long posited its evolutionary benefits. As a species, we have benefited from feeling anxious: %about possible future negative outcomes: 
for example, building safe dwellings to avoid being attacked by wild animals and poisonous insects, being respectful and kind to others to avoid social exclusion, and so on \cite{bateson2011anxiety}.
Yet, there is much we do not know such as the mechanisms of how anxiety develops %and manifests 
in our bodies and mind, how normal and clinical anxiety manifest in language, and effective mechanisms to cope with excessive anxiety. Further, %some have argued that 
the dramatic technological advances over the last few decades have rapidly and substantially changed our environments to such an extent (for instance, the widespread social media usage and human--technology interactions) that there is now a considerable mismatch between the current environment and what our anxiety response slowly evolved to address over millennia \cite{blease2015too,salecl2004anxiety,qureshi2022anxiety}. 
This has led some to argue that we are in an `age of anxiety' \cite{parish2000age,blease2015too}.

Anxiety has always been a part of normal human life. Some might feel anxious occasionally and in particular circumstances (state anxiety), and some might have a more diffuse sense of general anxiety over long periods of time (trait anxiety). 
Some of the major sources of anxiety include tests, sports, technology, and social anxiety \cite{zeidner2010anxiety}. More recently researchers have also identified a growing amount of climate change anxiety \cite{clayton2020climate}.
Mental health practitioners clinically diagnose anxiety disorders when the sense of anxiety is so high or so frequent that it impacts the day-to-day functioning of an individual. In the US alone, 6.8 million adults (3.1\% of the population) are believed to have Generalized Anxiety Disorder (GAD), 15 million adults (7.1\%) have Social Anxiety Disorder (SAD), 7.7 million adults (3.6\%) suffer from Post-traumatic Stress Disorder (PTSD), and everyone experiences stress from time to time.\footnote{https://adaa.org/understanding-anxiety/facts-statistics}
Anxiety disorders are highly co-morbid with depression \cite{kalin2020critical}, and together they constitute the \textit{internalizing disorders}.

% Trait, State
% General, Contextualized
% 

In addition to impacting well-being and happiness, anxiety (both normal and clinical) has been shown to impact cognitive outcomes such as learning, accurately performing tasks, productivity, as well as our physical bodies \cite{kim2005psychobiology}.
In contrast, calmness (being in a state of peace and tranquility) has been shown to improve health and cognitive outcomes \cite{siddaway2018reconceptualizing}.
\newcite{siddaway2018reconceptualizing} also show the importance of conceptualizing anxiety as a continuum from high calmness to high anxiety; and that both promoting calmness and decreasing excessive anxiety are important for overall health and well-being.
Thus, anxiety and calmness % understanding and coping with anxiety (and developing a sense of calmness) 
are active areas of research in Psychology, Medicine, Public Health, Social Science, Economics, and even the Humanities. %\footnote{Notable newer sub-areas of focus are anxieties associated with social media use, cyber bullying/harassment, dealing with hate speech, and climate change anxiety.} 

As is the case with many emotions, people express anxiety through language (consciously and unconsciously). In this work, we study how anxiety and calmness are associated with word meaning. We asked over a thousand respondents to rate 44,450 English words on a scale from $-3$ (associated with high calmness) to $+3$ (associated with high anxiety). 
% Each word was rated by 8.7 raters, on average.
We aggregated the responses to create a word--anxiety association lexicon that has a real-valued anxiety score for each of the words. We refer to this resource by the name \textit{WorryWords}. 
We show that the scores in WorryWords are highly reliable (a split-half reliability score of 0.82, which is in-line with scores for existing resources for other emotions). We use WorryWords to study how anxiety is related to other emotion constructs (valence, arousal, dominance, fear, anger, sadness, etc.). We show the rate at which anxiety and calmness words are acquired with age. Finally, we show that WorryWords can be used with simple word-counting methods to accurately track anxiety in text streams.
% (predicted anxiety arcs have average correlations of over 0.9 with the gold anxiety arcs).

We make WorryWords, and its translations in over 100 languages, freely available for research through the project home page.\footnote{\url{http://saifmohammad.com/worrywords.html}} 
Some areas where we envisage its use include:\\[-10pt]
\begin{compactenum}
\item  Understanding anxiety and the underlying mechanisms; how anxiety relates to other emotions; how it relates to our body; how anxiety changes with age, socio-economic status, weather, green spaces, etc. 
\item Determining how anxiety manifests in language; how language shapes anxiety; how culture shapes the language of anxiety; etc.
\item Tracking the degree of anxiety towards targets of interest such as climate change, %BB commercial products, 
government policies, biological vectors, etc. %; tracking common targets of anxiety;  
\item Identifying effective coping mechanisms and clinical interventions to manage anxiety.
\item Developing automatic systems for detecting anxiety; developing chat systems that are sensitive to nuances and diverse expressions of anxiety by people from various demographics.
\item Studying anxiety and uneasiness in story telling; its relationship with central elements of narratology such as conflict and resilience.
\item Studying how anxiety impacts social behaviour in physical and virtual environments.\\[-10pt]
\end{compactenum}
\noindent We also created an extensive list of ethical considerations (Section 9) that must be taken into consideration when using WorryWords. %BB (such as common pitfalls and best-practises when drawing conclusions from the use of the lexicon on one's data). 

% It also includes a list of expressly prohibited uses such as generate anxiety-inducing text for malicious purposes and drawing conclusions about an individual's feelings simply from what they may have said -- context matters. 

\section{Related Work}

%\noindent {\bf Emotion Constructs and Lexicons:} 
% Emotions are central to our lives, % interactions with the world, 
% motivations, decision making, and well-being. Yet, there is much we do not know about them, especially in terms of how we contruct them and how they are related to each other. 
It is generally agreed that the three primary dimensions of affect (our sense of feeling) are valence (positiveness--negativeness / pleasure--displeasure), arousal (active--passive), and dominance (dominant--submissive)  \cite{russell1980circumplex,Osgood1957}. 
There is also work showing that certain categorical emotions such as joy, anger, and fear are particularly important \cite{Ekman92,Plutchik80}. More recent work shows that we construct and label these emotions (a cognitive process) in our minds from various interoceptive signals (internal signals from our body to the brain) \cite{barrett2017theory}.

Anxiety is often considered a low-valence, high-arousal, and low-dominance emotion, but there is little work empirically exploring the degree to which it is correlated with these affective dimensions. Anxiety is also often compared to fear, but unlike fear which generally tends to be a reaction to a threat from the \textit{present} (something that has already manifested), anxiety is often associated with the \textit{possible} negative outcome of a \textit{future} event \cite{bateson2011anxiety}. 
Anxiety and stress are sometimes used synonymously and they are generally agreed to be very close emotions (both impact mind and body through excessive worrying, %BB uneasiness, 
loss of sleep, etc.), although stress is often associated with some physical stimulus and stress can be one of the causes of anxiety. %BB\footnote{See this article and infographic by NIH: https://www.nimh.nih.gov/health/publications/so-stressed-out-fact-sheet}

Manually created word--emotion association lexicons exist for valence, arousal, and dominance (\newcite{warriner2013norms} and \newcite{mohammad-2018-obtaining} for English;  \newcite{moors2013norms} for Dutch, and  \newcite{vo2009berlin} for German). % and by \newcite{redondo2007spanish} for Spanish).
The largest among these is the NRC VAD lexicon \cite{mohammad-2018-obtaining,vad-v2}: version 1 has entries for over 20,000 English words, and version 2 for over 44,000 unigrams and 10,000 bigrams (two-word sequences).
The NRC Emotion Lexicon \cite{MohammadT10} is a popular lexicon for categorical emotions; it has entries for over 14,000 English words. 

We do not know of existing large lexicons for anxiety; however, there exist some datasets of social media \textit{posts} manually annoatated with anxiety or stress labels. \newcite{rastogi2022stress} collected and annotated about 5,488 Reddit posts %and 10,000 Tweets 
with the labels 0 (no stress) and 1 (stress).  
\newcite{mitrovic-etal-2024-comparing} collected a series of questions and answers from Quora and Reddit, where the question contained the term `anxiety'.
\newcite{turcan-mckeown-2019-dreaddit} created a dataset of 4000 5-sentence segments from Reddit annotated for presence or absence of stress.
These social media datasets were compiled primarily  to train and test automatic systems that attempt to detect anxiety from text. For details on automatic anxiety detection, we refer the interested readers to \newcite{mitrovic-etal-2024-comparing,shen-rudzicz-2017-detecting,rastogi2022stress,winata2018attention}.
More recent work interviewed volunteers and analyzed their responses to identify language features such as the use of personal pronouns, negations, and emotion words with anxiety \cite{stade2023depression}.

As part of the WorryWords Project, we created a large word--anxiety association lexicon with a focus on reliable annotations.\footnote{In follow on work, we also compiled anxiety association norms for over 10K English phrases \cite{ww-2}.}
% (larger even than existing lexicons for VAD and categorical emotions such as anger and fear). 
We used existing emotion lexicons to examine how anxiety is related to other emotions. We also made use of a the \newcite{rastogi2022stress} dataset to show the usefulness of WorryWords in tracking anxiety in text streams. %how accurate anxiety arcs can be generated using WorryWords.

% \noindent {\bf NLP Work on Anxiety:} 

\section{Obtaining Human Ratings of Anxiety}

The keys steps in creating the anxiety dataset were: 
\begin{compactenum}
\item selecting the terms to be annotated 
\item developing the questionnaire
\item developing measures for quality control (QC)
\item annotating terms on a crowdsource platform
\item discarding data from outlier annotators (QC)
\item aggregating data from multiple annotators to determine the anxiety association scores
% for each of the terms. 
\end{compactenum}
We describe each of these steps below.\\[3pt]
\noindent {\bf 1. Term Selection.}
% Our goal was to annotate a large number of common English terms. 
To allow for the study of anxiety with respect to other affectual and linguistically interesting properties, we included terms from: % various existing resources: 
the NRC Emotion Lexicon \cite{MohammadT13,MohammadT10}, the NRC VAD Lexicon \cite{mohammad-2018-obtaining}, and the terms from the Prevalence dataset \cite{brysbaert2019word} 
%(with manually obtained labels for how widely each term is known) 
(specifically, words marked as being known to over 70\% of the respondents). Summary details of each of these is provided in the Appendix.
The union of the above sets resulted in 44,450 terms that were then annotated for association with anxiety and calmness. % as described below.

\noindent{\bf 2. Anxiety Questionnaire.}
The questionnaires used to annotate the data 
 were developed  across two rounds of 
 pilot annotations. 
 Detailed directions, including notes directing them to consider predominant word sense (in case the word is ambiguous) and example questions (with suitable responses) were provided. (See Appendix.)
 The primary instruction and the question presented to annotators is shown below.\\[-17pt]
% The annotation questions and the instructions for the annotators are shown in a supplementary file. 

{
\noindent\makebox[\linewidth]{\rule{0.48\textwidth}{0.4pt}}\\% [-8pt]
{ \small
% Summary Instructions
\noindent Consider anxiety to be a broad category that includes:\\
\indent \textit{jittery, antsy, insecure, nervous, unease, tense, worried,\\ 
\indent unnerving, nerve-racking, apprehensive, fretful, troubled,\\ 
 \indent self-doubting, discontented, concerned, and keyed up}\\
Consider calmness to be a broad category that includes:\\
\indent \textit{calm, relaxed, comforted, serene, at ease, self-assured, \\ 
\indent carefree, composed, collected, untroubled, peaceful,\\
\indent contented, unconcerned, indifferent, and uninvolved}
% nonchalant, uninterested, 
% If you do not know the meaning of a word or are unsure, you can look it up in a dictionary (e.g., the Merriam Webster) or on the internet.

% Quality Control

% Some questions have pre-determined correct answers. If you mark these questions incorrectly, we will give you immediate feedback in a pop-up box. An occasional misanswer is okay. However, if the rate of misanswering is high (e.g., >20\%), then all of one's HITs may be rejected

\noindent Select the options that most English speakers will agree with.\\[4pt]
\noindent Q1.  <term> is often associated with feeling:\\[-1pt]
\indent 3: very anxious \hspace{14mm} -1: slightly calm\\[-1pt]
\indent 2: moderately anxious \hspace{5mm}  -2: moderately calm\\[-1pt]
\indent 1: slightly anxious \hspace{10mm}  -3: very calm\\[-1pt]
\indent 0: not associated with feeling anxious or calm\\[-8pt]
% \indent -1: slightly calm\\
% \indent -2: moderately calm\\
% \indent -3: very calm\\[-8pt]
}
\noindent\makebox[\linewidth]{\rule{0.48\textwidth}{0.4pt}}\\[-12pt]

}
% \noindent The full questionnaire will be made available on the project webpage. 

\noindent{\bf 3. Quality Control Measures.}
About 2\% of the data was annotated beforehand by the authors and interspersed with the rest. These questions are referred to as \textit{gold} (aka \textit{control}) questions. 
% During crowd annotation, 
% We interspersed the gold questions with the other questions.
% and the annotator is not aware which is which. However, 

Half of the gold questions were used to provide immediate feedback to the annotator (in the form of a pop-up on the screen) in case they mark them incorrectly. We refer to these as \textit{popup gold}. This helps prevent the situation where one annotates a large number of instances without realizing that they are doing so incorrectly. 
It is possible, 
that some annotators share answers to gold questions with each other (despite this being against the terms of annotation). 
% The gold questions also served as examples to guide the annotators.
Thus, the other half of the gold questions were also separately used to track how well an annotator was doing the task, but for these gold questions no popup was displayed in case of errors. 
We refer to these as 
\textit{no-popup gold}.\\[-13pt]
% If a crowd worker answered a gold question incorrectly, then they were immediately notified.
% \sm{the annotation was discarded, and an additional annotation was requested from a different annotator}. 

\noindent{\bf 4. Crowdsourcing.} 
We setup the anxiety annotation task on the crowdsourcing platform, {\it Mechanical Turk}.
 In the task settings, we specified that we needed annotations from ten people for each word.
We obtained annotations from native speakers of English residing around the world. Annotators were free to provide responses to as many terms as they wished. 
The annotation task was approved by 
% the National Research Council Canada's Institutional 
our institution's review board.
% , which reviewed the proposed methods to ensure that they were ethical.
% We used CrowdFlower's gold annotations scheme for quality control, wherein 

\noindent {\it Demographics:} About 87\% of the respondents who annotated the words live in USA. The rest were from India, United Kingdom, and Canada. 
The average age of the respondents was 38.9 years. Among those that disclosed their gender, about 54\% were female and about 45\% were male.\footnote{Respondents were shown optional text boxes to disclose their demographic information as they choose; especially important for social constructs such as gender, in order to give agency to the respondents and to avoid binary language.}\\[-13pt]

\noindent{\bf 5. Filtering.} % and Re-annotation} %  as part of Quality Control} 
If an annotator's accuracy on the gold questions (popup or non-popup) fell below 80\%, then they were refused further annotation, 
 and all of their annotations were discarded (despite being paid for).
% We then obtained fresh annotations for those terms from other annotators.
% This served as a mechanism to avoid malicious and random annotations.
  % However, because of the way the gold questions work in CrowdFlower, they were annotated by more than six people. 
%  Both the minimum and the median number of annotations per item was 10. 
 See Table~\ref{tab:ann} for summary statistics.\\[-13pt]

\noindent{\bf 6. Aggregation.} 
Every response was mapped to an integer from -3 (very calm) to 3 (very anxious). % as follows: 
% \begin{compactitem}
%     \item high anxiety: 3
%     \item moderate anxiety: 2
%     \item slight anxiety: 1
%     \item neither anxiety nor calmness: 0
%     \item slight calmness: -1
%     \item moderate calmness: -2
%     \item high calmness: -3
% \end{compactitem}
The final anxiety score for each term is simply the average score it received from each of the annotators.
% The scores were then linearly transformed to the interval: -1 (highest calmness) 
% to 1 (highest anxiety).
% Since degree of emotion is a unipolar scale, 
% We linearly transform the -1 to 1 scores to scores in the range 0 (least emotion intensity) to 1 (the most emotion intensity).
We also created a categorical version of the lexicon by labeling all words that got a score $\geq 2.5$ as associated with \textit{high anxiety}, $\geq 1.5$ and $< 2.5$ as \textit{moderate anxiety}, $\geq 0.5$ and $< 1.5$ as \textit{slight anxiety},
$> -0.5$ and $< 0.5$ as \textit{neither anxiety nor calmness}, and so on.  
We refer to the
list of words
along with their 
 real-valued 
final scores and categorical labels as the {\it WorryWords Lexicon}. 

\begin{figure}[t]
	     \centering
	     \includegraphics[width=0.45\textwidth]{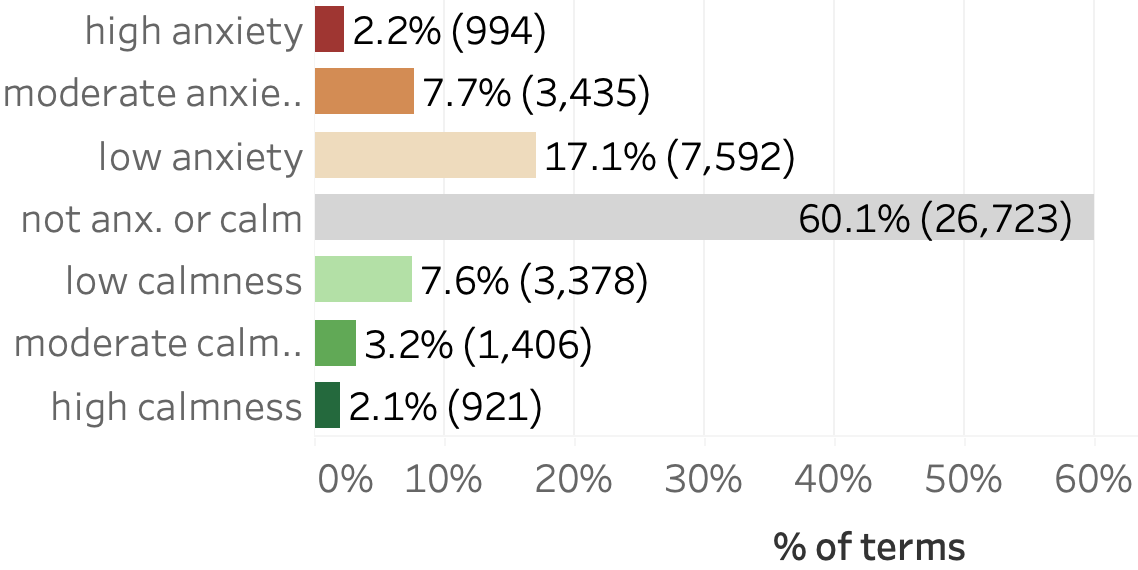}
	    \vspace*{-1mm}
      \caption{Distribution of terms in WorryWords: percentage of terms associated with each class. (The total number is shown in parenthesis.)}
	     \label{fig:WorryWords-distrib}
\vspace*{-3mm}
	 \end{figure}

Table \ref{tab:examples} in the Appendix shows example entries from the lexicon. % with the highest and lowest scores.
Figure \ref{fig:WorryWords-distrib} shows the distribution of the different classes. As expected, most words are associated with neither anxiety nor calmness ($\sim$60\%), but it is worth noting that about 27\% of the words are associated with anxiety (to some degree) and about 13\% of the words are associated with calmness (to some degree).

\begin{table*}[t!]
\centering
% \begin{center}
%\vspace*{-4mm}
\small{
\begin{tabular}{lrrrrrrrrr}
\hline 
%       & \multicolumn{2}{c}{\bf \#instances from} &\\
%				 & Tweets-2017 & Tweets-2018 &\\

{\bf Dataset} 	& \bf \#words	&\bf Annotators   & \bf \#Annotators &\bf \#Annotations &\bf MAI  &\bf SHR ($\rho$)		 &\bf SHR ($r$) \\\hline
unigrams 		& 44,450 &US, India, UK, Canada  & 1,020 & 375,796  & 8.5 	& 0.82 &0.89\\
% bigrams 		& 10,000 &worldwide  & 1,020 & 10 	& 243,295\\
%\hline
%\bf Total  &54,000 		 	& & & &  \bf 778,085\\
 \hline
\end{tabular}
}
\vspace*{-1mm}
\caption{\label{tab:ann} {A summary of the WorryWords annotations. MAI = mean  annotations per word. SHR, measured through both Spearman rank and Pearson's correlations (last two columns), indicate high reliability.}
}
% \vspace*{-3mm}
% \end{center}
\end{table*}

\section{Reliability of the Annotations} 
A useful measure of quality is the reproducibility of the end result---repeated independent manual
annotations from multiple respondents should result in similar  scores.
To assess this reproducibility, we calculate
average {\it split-half reliability (SHR)} over 1000 trials.\footnote{SHR is a common way to determine reliability of responses to generate scores on an ordinal scale %in the fields of psychology and psycholinguistics 
\cite{weir2005quantifying}.} 
% SHR a commonly used approach to
% determine consistency in psychological studies, that we employ as follows.   
All annotations for an
item are randomly split into two halves. Two separate sets of scores are aggregated, just as described in Section 3 (bullet 6), from the two halves. 
% but independently from the two SHR halves.  
% Then the correlation between the two sets of scores is calculated. 
Then we determine how close the two sets of scores are (using a metric of correlation). % using Spearman Rank Correlation. 
This is repeated 1000 times and the correlations are averaged.
The last two columns in Table~\ref{tab:ann} show the results (split half-reliabilities). Scores of 0.82 (Spearman rank correlation) and 0.89 (Pearson correlation) indicate high reliability of the real-valued scores obtained from the annotations. (For reference, if the annotations were random, then repeat annotations would have led to an SHR of 0. Perfectly consistent repeated annotations lead to an SHR of 1.) 
%  Past emotion lexicon work such as the NRC Emotion Intensity Lexicon \cite{} report SHRs in the low 0.9s with 32 annotations per item.)

If the real-valued scores are converted to discrete categories (for example, into 3, 7, or 10, classes), then we can make use of another measure of reliability that factors in the classes.
% % A key aspect here is the level of granularity of the final lexicon. 
For example, if we just want three classes: \textit{associated with anxiety, associated with calmness,} and \textit{neither}. Then one could simply split the -3 to 3 range into three equal partitions of -3 to -1 (calmness), -1 to 1 (neither), and 1 to 3 (anxiousness). % \footnote{Individual classes can also be determined from ranges that are not equal sized. For this analysis we use equal-sized partitions for simplicity.}
To determine reliability of the discrete labels obtained from annotations, we first randomly split all annotations into two halves (just as before).
Now, if scores for a term \textit{x} obtained from the two halves place \textit{x} in different classes (e.g., if the two scores are 0.36 and 1.72), then this contributes to a lower overall reliability.
In contrast, even if the scores for \textit{x} from the two halves vary a bit, but both scores place \textit{x} in the same class (e.g., if the two scores are 1.82 and 1.79) then this contributes to a higher overall reliability.
Thus we can compute reliability from the extent to which the two halves of annotations result in the same discrete class. We will refer to this as the \textit{Split-Half Class Match Percentage (SHCMP)}.

Note, however, that using discrete classes means that many items that have similar real-valued scores from two independent sets of annotations may be on different sides of a class boundary. Thus even if the original annotations are reliable, using discrete classes adds some unreliability inherently. To get a sense of this, we also calculate the degree to which the split-half scores are close to each other (without regard to class boundaries): e.g., the percentage of times two  independent sets of annotations for an item lead to scores that are closer than a pre-chosen threshold (0.1, 0.25, etc.). We will call this the \textit{Split-Half Closeness Percentage (SHCloseP)}.

% If we are interested in a much finer granularity, say corresponding to 10 bins of size 0.2 from -1 to 1, then the scores from the two SHR halves must be markedly close to each other for most items to be placed in the same class.

% For example, if we just want three classes: associated with anxiety, associated with calmness, and neither. Then one could simply partition the -1 to 1 range into three equal subranges of -1 to -0.33 (calmness), -0.33 to 0.33 (neither), and 0.33 to 1 (anxiousness).
% In this case, if scores for a term \textit{x} obtained from the two SHR halves place \textit{x} in different classes (e.g., if the two scores are 0.26 and 0.47) then this contributes to a lower overall reliability.
% In contrast, even if the scores for \textit{x} from the two SHR halves vary a bit, but both scores place \textit{x} in the same class (e.g., if the two scores are 0.82 and 0.87) then this contributes to a higher overall reliability.

Table \ref{tab:shr}  shows the SHCMP and SHCloseP for the WorryWords annotations for various levels of granularity (\#classes). The second column shows the class size% corresponding to the \#classes
---obtained by dividing the -3 to 3 range equally by \#classes.\footnote{SHCMP can also be determined from custom ranges that are not of the same size. Here we show SHCMP with equal size bins for simplicity.} 
The third column shows the expected SHCMP when terms are given scores at random.
The fourth column shows SHCMP when using the actual WorryWords annotations. 
The fifth column shows SHCloseP when using different bin sizes as threshold. 
% The SHCMP scores act as measures of SHR in case of discrete bins.

% \begin{table}[t]
% \small
%     \centering
%     \begin{tabular}{r c c c c}
%     \hline
%                         & & \textbf{Random}    &\multicolumn{2}{c}{\textbf{WorryWords}}\\
%   \textbf{\#Bins}    &\textbf{Bin size} &\textbf{SHCMP} \% & \textbf{SHCMP\%} & \textbf{SD} \\
%      \hline
% 	10 &0.60    &10.0   &84.9     &0.10\\
%     7 &0.86    &14.3   &91.6     &0.10\\
% 	5  &1.20	   &20.0   &96.8     &0.07\\
% 	4  &1.50   &25.0   &98.4     &0.04\\
% 	3  &2.00   &33.0   &99.4     &0.03\\
% 	2  &3.00    &50.0   &100.0    &0.01\\
%         \hline
%     \end{tabular}
%     \caption{Split-Half Class Match Percentage (SHCMP) 
%     % in ordinal classes of different bin sizes 
%     obtained from the anxiety annotations when
%     splitting the -3 to 3 score range into equal-sized bins (partitions) pertaining to ordinal classes. SD: standard deviation.}
%     \label{tab:shr}
%      \vspace*{-3mm}
% \end{table}

\begin{table}[t]
% \small
\resizebox{0.48\textwidth}{!}{
 %   \centering
 \begin{tabular}{r c c c c}
    \hline
                        & & \textbf{Random}    &\multicolumn{2}{c}{\textbf{WorryWords}}\\
  \textbf{\#Bins}    &\textbf{Bin size} &\textbf{SHCMP} & \textbf{SHCMP} & \textbf{SHCloseP} \\
     \hline
	10 &0.60    &10.0   &0.62  &84.9 \\
    7 &0.86    &14.3    &0.71  &91.6 \\
	5  &1.20   &20.0    &0.78  &96.8 \\
	4  &1.50   &25.0    &0.79  &98.4 \\
	3  &2.00   &33.0    &0.87  &99.4 \\
	2  &3.00    &50.0   &0.89  &100.0 \\
        \hline
    \end{tabular}
    }
    \caption{Split-Half Class Match Percentage (SHCMP) and Split-Half Closeness
    Percentage (SHCloseP)
    % in ordinal classes of different bin sizes 
    obtained from the anxiety annotations when
    splitting the -3 to 3 score range into equal-sized bins (partitions) pertaining to ordinal classes. SD: standard deviation.}
    \label{tab:shr}
     \vspace*{-3mm}
\end{table}

% \noindent \textsc{\bf Summary of Main Results:} SHR scores of 0.9 show  that highly reliable 7-class word--anxiety ratings of association can be obtained through crowdsourcing. 

Observe that the SHCMP scores are markedly higher than what one would get with just random annotations. Further, SHCMP scores are fairly high in case of a small number of discrete classes (bins), but drops off markedly as the number of classes is increased. Note the considerably higher SHCloseP scores which indicate that much of the unreliability in discrete classes is due to items falling on either side of class boundaries, and that when there is a class mismatch the involved classes are close to each other (smaller error) as opposed to distant classes (bigger error).

% of over 95\% for 5 and fewer classes indicate very high levels of reliability.

%      \textbf{Speaker} & \multicolumn{2}{c|}{\textbf{Valence}} & \multicolumn{2}{c|}{\textbf{Arousal}} & \multicolumn{2}{c|}{\textbf{Dominance}} \\

\section{WorryWords: Research \& Applications}

WorryWords has a wide variety of uses and applications as listed in the Introduction. 
We explore:
% Since this paper introduces the lexicon, 
%We include here:
\begin{compactitem}
    \item Two diverse examples where lexicographic analysis of the words helps answer research questions about anxiety (Section 6).
    \item An example (with evaluation) of how the WorryWords can be used to accurately track anxiety in text streams (Section 7). % Accurately tracking the flow of anxiety (at an aggregate level) 
    This has both commercial and research applications.
\end{compactitem}

% \section{Lexicographic Analysis of WorryWords}

% Lexical Analysis of WorryWords to Shed Light on
\section{Using WorryWords to Answer Research Questions about Anxiety}

\subsection{From a lexical standpoint, how is anxiety related to other emotions?}

% \@ VAD Dimensions and\\ Anxiety vs.\@ Discrete Emotions}
An important question in understanding any emotion is how it relates to other emotions.
With the creation of the first large-scale lexicon of word--anxiety associations,
we can now study how these norms correlate with other key emotion constructs such as
valence (V), arousal (A), and dominance (D), as well as categorical emotions such as sadness, fear, and anger. Since anxiety is commonly understood to be a negative (low-valence emotion), we expect it to have a negative correlation with valence (negative situations are often associated with anxiety). However, we have no prior evidence to indicate how strong this negative correlation is expected to be; especially since, there can be many term pairs w1 and w2 such that even though w1 has lower valence than w2, w2 is associated with a higher degree of anxiety than w1. For example, 
\textit{hopeless} and \textit{frenzy}, \textit{lethargic} and \textit{public speaking}, \textit{weak} and \textit{test}, etc.

Similarly, we expect anxiety to have a positive correlation with arousal (since high anxiety terms are also expected to be high arousal terms), and a negative correlation with dominance (since higher anxiety tends to co-occur with situations of loss of control). Yet, again, it is unclear how strong these correlations truly are across a vast vocabulary of words.
Understanding these relationships has implications in psycholinguistics and lexical semantics: e.g., whether anxiety moderates word recognition even when controlling for valence. Further, in the context of lexicon-creation: If the magnitudes of these correlations are very high, then one can simply take an existing VAD lexicon, and keep only the low valence, high arousal, and/or low dominance terms to create an anxiety lexicon. 

\noindent {\bf VAD:} The first row (and also the first column) of Table \ref{tab:corrS-anx-vad} shows the correlations of anxiety association scores of terms in WorryWords with the valence, arousal, and dominance scores of terms in the NRC VAD Lexicon v2 \cite{vad-v2}.
Observe that anxiety has a moderate negative correlation with valence, a mild correlation with arousal, and a mild negative correlation with dominance.
For additional context, the bottom 3 rows (and last 3 columns) of Table \ref{tab:corrS-anx-vad} show the pair-wise correlations between valence, arousal, and dominance.  
It shows for example, that valence and dominance tend to be more correlated than any of the other pairs (a result also seen in other VAD lexicons such as the Warriner Lexicon \shortcite{warriner2013norms});
and in terms of the magnitude, the degree of (negative) correlation between valence and anxiety is lower than that of valence and dominance.

\begin{table}[t]
\small
    \centering
    \begin{tabular}{lrrrr}
        \hline
                 &\textbf{anxiety} & \textbf{valence} & \textbf{arousal} & \textbf{domin.} \\
     \hline
\textbf{anxiety}	&1.000		&-0.471		&0.237		  &-0.218	\\	
\textbf{valence}	&-0.471		&1.000		&-0.014		  &0.568	\\
\textbf{arousal}	&0.237		&-0.014		&1.000		  &0.286	\\
\textbf{domin.}	&-0.218		&0.568		&0.286	      &1.000\\
        \hline
    \end{tabular}
    \vspace*{-2mm}
    \caption{Spearman rank correlations of association scores between various emotion pairs, for ~44k terms.}
    \label{tab:corrS-anx-vad}
    \vspace*{-4mm}
\end{table}

\begin{table*}[t]
\small
    \centering
    \begin{tabular}{lrrrrrrrrr}
              \hline
	&\textbf{anxiety} &\textbf{anger}	&\textbf{anticipn.}		&\textbf{disgust}	&\textbf{fear}	&\textbf{joy}	&\textbf{sadness}	&\textbf{surprise}	&\textbf{trust}\\
     \hline
\textbf{anxiety}	&1.000		&0.336		&-0.011		&0.279		&0.357		&-0.132		&0.272		&0.082		&-0.137		\\
        \hline
    \end{tabular}
    \caption{Spearman rank correlation between the word--emotion association scores for anxiety and 8 other emotions.} 
    % For a given anxiety--emotion pair, the correlation is computed for terms from the NRC Emotion Intensity Lexicon that are marked as having some non-zero association with the emotion.}
    \label{tab:corrP-anx-8emo-small}
\end{table*}

% To gain further insights on the relationship of anxiety with VAD, 
We also created an interactive visualization that shows the WorryWords terms on valence--arousal and valence--dominance spaces (Figure \ref{fig:anx-scat-vad} in the Appendix shows a screenshot). (Hovering over individual points shows the corresponding term as well as anxiety, V, A, and D scores.)
We observe that, as expected, a majority of the anxiety words (score $>$ 0.5) occur in the top-left quadrant of the V--A space: low V and high A. Nonetheless, there are still non-negligible number of terms that occur in the top-right quadrant (such as \textit{adrenalin}, \textit{revolutionist}, and {\it whizzing}) and in the bottom-left quadrant (such as \textit{joylessness}, \textit{disheartenment}, and \textit{disreputableness}). Comparatively much fewer anxiety terms occur in the bottom right (high V and low A) quadrants. Examples include \textit{seatbelt} and \textit{postoperative}. In contrast, the majority of the calmness terms (score $<$ $-$0.5) occur in the high V quadrants. Examples, of low V terms associated with calmness include \textit{pity} and \textit{aloof}.
The V--D plots show that anxiety terms tend to be concentrated in the bottom-left quadrant, whereas calmness terms are most commonly found in the top-right quadrant. Anxiety terms associated with high D include terms such as \textit{skyrocket}, \textit{canny}, \textit{gunfight}, and \textit{tyrant}; anxiety terms associated with low D include \textit{persecuted} and \textit{gutlessness}.
%BB some errors also; some terms that can be seen as pos or negative
%BB V-D quadrant examples

% Figure \ref{fig:anx-scat-vad} e and f show anxiety and calmness terms in the V--A and V--D spaces respectively. We can see that the majority of the anxiety terms occur in the low-dominance and low-valence quadrants whereas the majority of the calmness terms occur in the high valence and high dominance quadrants.

\noindent {\bf Categorical Emotions:} Table \ref{tab:corrP-anx-8emo-small} shows the correlations of anxiety 
% association 
scores of terms in WorryWords with the anger, anticipation, disgust, fear, joy, sadness, surprise, and trust scores of terms in the NRC Emotion Intensity Lexicon v1 \cite{LREC18-AIL}. Observe that anxiety seems to essentially have no correlation with anticipation, joy, and surprise (correlation magnitudes less than 0.2); and low correlation with the negative emotions: anger, disgust, fear, and sadness. Among these emotions, anxiety has the highest correlation with fear.

\noindent \textit{Discussion:} Overall, we see that while anxiety associations follow some expected trends such as pertaining to low valence, the magnitudes of correlations are moderate with valence and low with arousal and dominance. Thus anxiety presents itself in many different ways and cannot be characterized simply with VAD.
The fact that anxiety has the highest correlation with fear (among the four negative categorical emotions) is in line with the belief that fear is a key aspect of anxiety \cite{rosen1998normal,shin2010neurocircuitry}.

\subsection{At what rate do children acquire words associated with anxiety? How does this change with the Age of Acquisition?}
% \subsection{How does the rate of acquiring anxiety-associated words change with age in children?} 

Feeling anxious is a normal part of childhood and growing up \cite{beesdo2009anxiety}. The Cleveland Clinic describes how very young children (6-month to 3-year olds) can have separation anxiety from parents and caregivers; pre-schoolers often develop phobias towards specific things such as insect, darkness, thunder, heights, etc.; and school age brings many sources of anxiety such as going to a new school, exams, social interactions, and managing school work.\footnote{https://my.clevelandclinic.org/health/diseases/anxiety-in-children}
Over the course of development, the way children express, experience, and communicate their emotions, including anxiety, changes (Bailen et al., 2019; Thompson, 1991). Understanding these changes is instrumental in promoting healthy socio-emotional functioning across all stages of development.
Thus, the associations of emotions with different ages is a key aspect studied by researchers in child development.

With WorryWords we can now look specifically at the relationship between anxiety-associated words and age. Specifically, in our work, we examine the correlation between the age at which words are acquired by children in their vocabulary (the age they learn the meanings of words) and the anxiety associated with those words.

\begin{figure}[t]
	     \centering
	     \includegraphics[height=0.3\textwidth, width=0.46\textwidth]{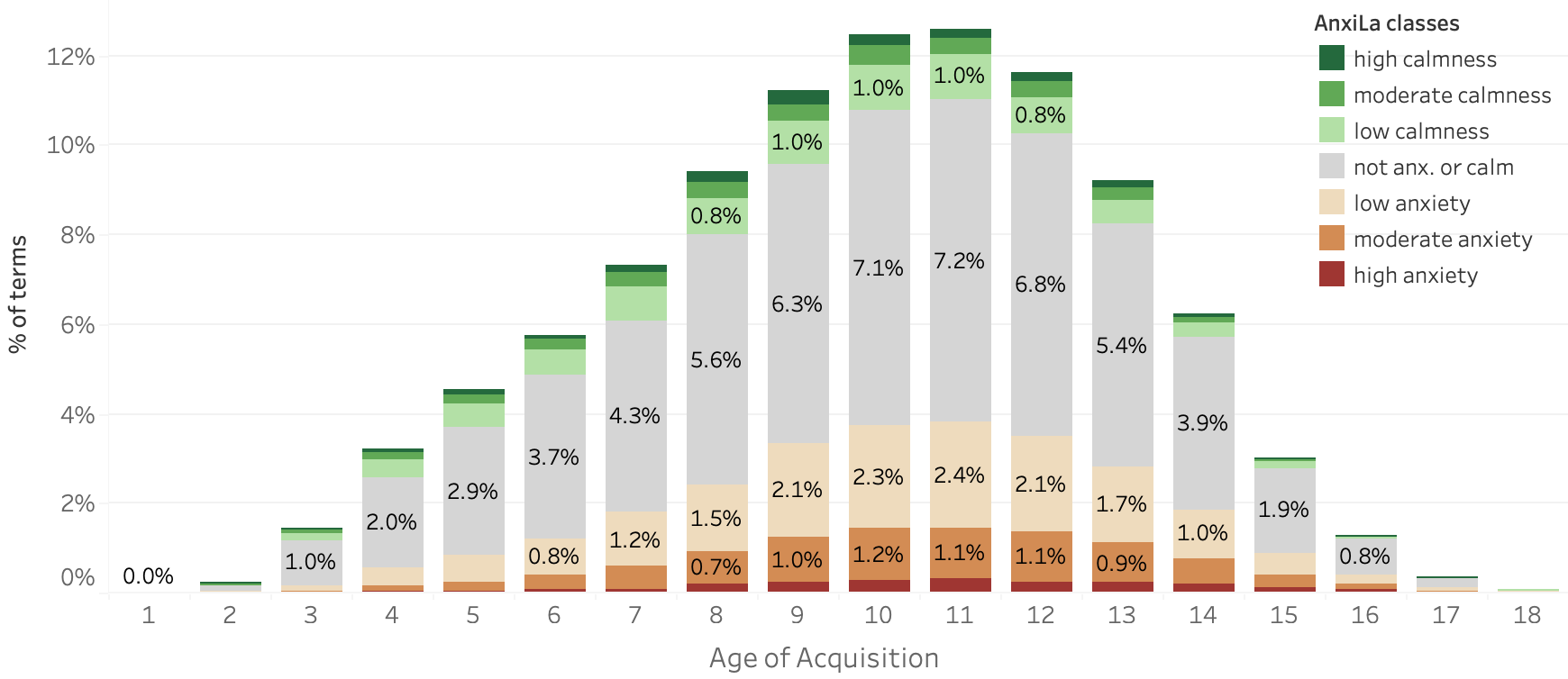}
      \vspace*{-2mm}
	     \caption{Percent of WorryWords terms by AoA.} % from age 1 to 18.}
	     \label{fig:anx-aoa}
      \vspace*{-3mm}
	 \end{figure}

We do so by examining the words with entries in both WorryWords and \newcite{kuperman2012age} age of acquisition dataset (which includes the age at which $\sim$30K English words are commonly acquired by children).
Figure \ref{fig:anx-aoa} shows the percentage of words in WorryWords that are acquired at various ages. Subsets of words 
% acquired within a year 
corresponding to the three anxiety classes are shown in shades of orange, corresponding to the three calmness classes are shown in shades of green, and the amount of words neither associated with anxiousness nor calmness are shown in grey.

Observe that the percentage of words acquired increases steadily with age, peaking at 11, and then decreasing rapidly after that. The percentage of anxiety and calmness words also largely follows the same pattern, but some notable additional observations can be made. At a very early age (1 to 4 years) the percentage of calmness words acquired is higher than the percentage of anxiety words. However, from age 5 onwards, children learn more anxiety words than calmness words. 

\begin{figure}[t]
	     \centering
	     \includegraphics[width=0.46\textwidth]{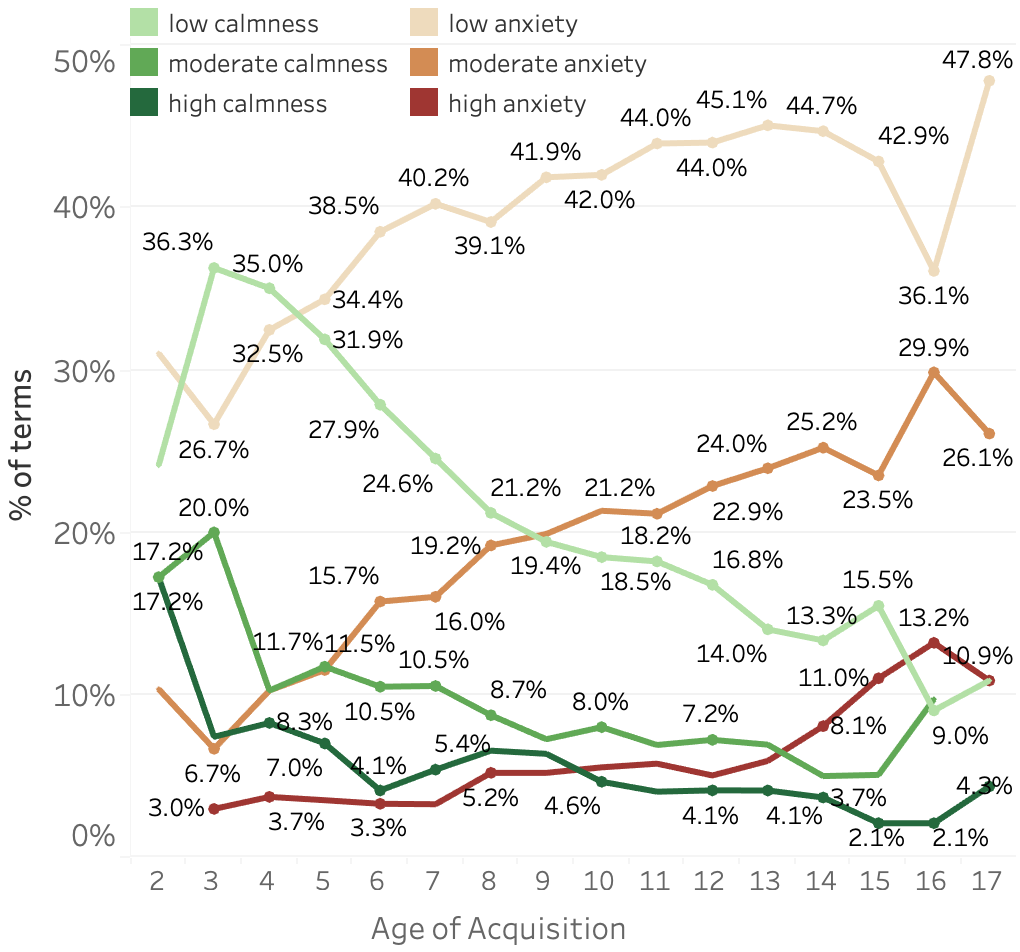}
      \caption{For all of the non-neutral WorryWords terms acquired at age \textit{x}: the percentage corresponding to each of the non-neutral anxiety--calmness bins. (The six percentages for each age sum up to 100\%.)}
	  %   \caption{For each non-neutral WorryWords bin \textit{x} and age \textit{y}: the percentage of \textit{x} terms acquired at \textit{y} w.r.t. all non-neutral terms acquired at \textit{y}.}
	     \label{fig:anx-aoa-2}
      \vspace*{-3mm}
	 \end{figure}

% To better understand the relative numbers of anxiety and calmness words acquired each year, 
Figure \ref{fig:anx-aoa-2} shows the percentage of words acquired each year for each of the anxiety and calmness classes; i.e., after disregarding the \textit{neither anxious nor calm} words, what percentage of words acquired in a year correspond to each of the six other classes. %(Also, since it shows the percentage of words acquired in each year, 
(For every x-axis value, the six percentages sum up to 100\%.) 
Observe that from age 3 onwards, the percentage of anxiety words increases steadily with age (for all three anxiety classes), whereas, the percentage of calmness words decreases with age (for all three calmness classes).

\noindent \textit{Discussion:} The notable trends in the relative rates of acquisition of anxiety and calmness terms with age have implications in developmental psychology and evolutionary linguistics \cite{price2013evolutionary,breggin2015biological}. They may also be relevant in the understanding of and coping with childhood anxieties \cite{wright2010depression,legerstee2010cognitive}. We hope that WorryWords will spur further research in these areas.

\section{Capturing Anxiety Arcs in Text Streams using WorryWords}

A common use of emotion lexicons is to determine emotion arcs and trends in how emotions change in a stream of textual data. For example, determining changes in the degree of anger towards a product in social media over time \cite{breeze2020angry}, tracking happiness and well-being of different cities and counties through geo-located posts on X (formerly Twitter) in public health research \cite{frank2013happiness}, generating emotion arcs of stories in digital humanities \cite{mohammad-2011-upon,ohman-etal-2024-emotionarcs}, etc. The lexicons may be used in combination with supervised systems and large language models (where such resources are available), or they may be used on their own where interpretability, low-carbon foot print, and simpler systems are valued --- which is often the case in psychology and public health research.
Lexicon-only methods are often markedly less accurate at determining emotions of \textit{individual sentences}. However, \newcite{teodorescu-mohammad-2023-evaluating} have shown that on the task of generating emotion arcs (where emotion scores of hundreds, if not thousands, of individual sentences are averaged to determine points on the emotion arc) lexicon-only methods are competitive with more sophisticated neural and large language models.%
% obtaining nearly identical correlations with gold emotion arcs on 43 dataset and 10 emotion categories.
% They argue that this is because when generating emotion arcs the goal is to accurately determine whether the average of the emotion scores of a \textbf{set of instances} is higher or lower than the average for another \textbf{set of instances}. If the sets include even just a few hundred instances, automatic methods can get these decisions right as often a majority of instances in a set have overt emotion-associated words.
\footnote{Even though an individual lexicon entry only provides the emotion association with the predominant sense (and does not take context into account), by tracking the emotions associated with hundreds of co-occurring words in a text stream %, the lexicon approach to generate arcs 
we make use of context. 
% Each bin acts as contextual text.
 If for example, an anger-inducing event occurs, then 
 the corresponding bin is likely include many words 
 % describing this context 
 that will have an association with anger.}
% A marked increase in anger-associated words this week compared to last week in tweet about a product, is a strong indicator that there is more anger this week than last.
In this section we explore:
\begin{compactenum}
    \item How accurate are anxiety arcs generated using WorryWords alone?
    \item Given the close relationship between anxiety and other emotion categories such as fear, anger, and low valence, how good are lexicons for other emotions at capturing anxiety arcs? 
\end{compactenum}
\noindent Together these results tell us about the usefulness of WorryWords in the task of generating emotion arcs, as well as how anxiety-associated and other-emotion-associated words present themselves in sentences that have been marked for anxiety.

\subsection{Creating the Anxiety Arcs}

First we quickly summarize how emotion arcs are created. The user (anyone interested in tracking anxiety for a text stream of interest) provides the:\\[-10pt]
\begin{compactitem}
    \item Text Stream: A series of instances (sentences, posts, etc.) in order of interest (e.g., posts on X that mention \textit{\#climatechange} ordered by timestamps, % (date and time of posting), 
    sentences from a novel).\\[-10pt]
    \item Bins: A pre-chosen grouping (day, month, year, \textit{k} sentences, etc. --- based on granularity  of interest)\footnote{For e.g., anxiety in tweets about climate change where one wants to track changes month by month, each bin includes the tweets posted in the corresponding {\it month}.}.
    An additional parameter relevant when the bin is specified in terms of number of sentences (say, \textit{k}) is the step size. For example, a step size of \textit{s} instances means that after including sentences 1 to \textit{k} in the first bin, the next bin includes instances $1+s$ to $1+s+k$, then $1+2s$ to $1+2s+k$, and so on. Commonly used step sizes are 1 (highly overlapping bins -- leading to smoother arcs) and \textit{k} (non-overlapping bins).\\[-10pt]    
\end{compactitem}
We generate the predicted arc for this text stream by simply taking the average anxiety association scores of words (taken from WorryWords) for each bin. Terms not found in the lexicon are disregarded when determining these scores. The gold arc can be created by taking the averages of the manually labeled anxiety scores (labels) of each of the posts in the bins.
The closer the predicted arc to the gold arc, the more accurate it is. 

\begin{figure*}[t]
	     \centering
	     \includegraphics[width=0.9\textwidth]{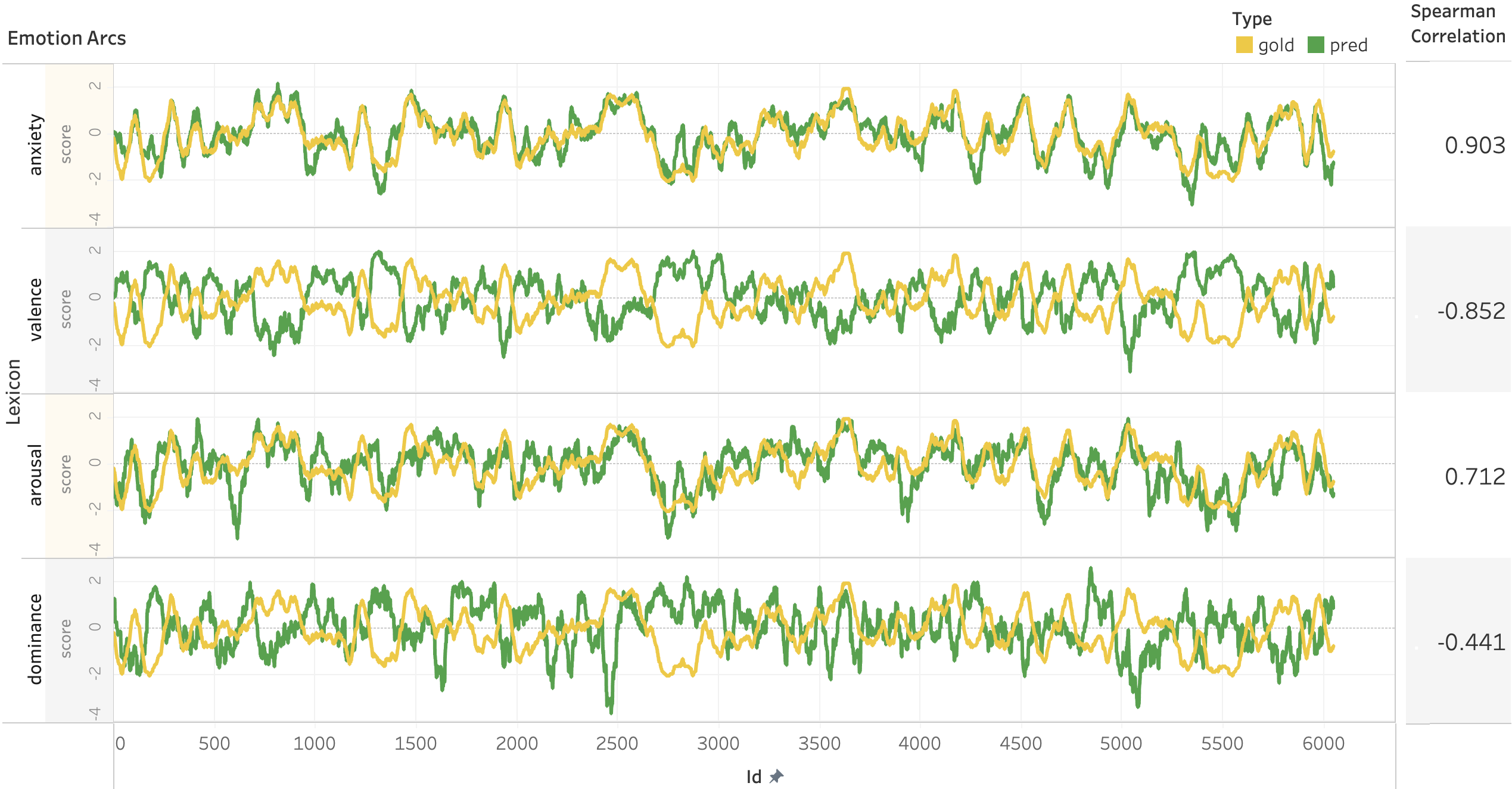}
      \vspace*{-3mm}
	     \caption{Predicted (green) anxiety arcs when applying four different types of association scores to one of the test data text streams: 
a. anxiety association from WorryWords, b. valence (from NRC VAD), c. arousal (from NRC VAD), and dominance (from NRC VAD). The gold (yellow) arc pertaining to the same text stream is repeated in all four figures for reference. Bin size used to create the arcs: 50 posts. On the right is the average correlation between 1000 pairs of gold and predicted arcs for each case.}
	     \label{fig:anxarc-avad}
\vspace*{-5mm}
	 \end{figure*}

However, there exists no cohesive text streams dataset with annotations for anxiety. That said, there exist datasets where independent sentences are annotated for anxiety, such as the stress dataset by \newcite{rastogi2022stress}. 
% \noindent {\bf Creating Anxiety Text Streams:} 
Since this dataset is not associated with a particular event where there might be ups and downs of emotions, a completely random ordering of the data or using timestamps to order the data into a text stream, produces of a flat line anxiety arc. 
% We generated 1000 different text streams (pertaining to various different complex gold anxiety arc shapes) from the \newcite{rastogi2022stress} dataset 
% to create to evaluate automatic methods of determining anxiety arcs.
% For any given pre-chosen window size, 
Thus, we created new text streams by
sampling posts from the \newcite{rastogi2022stress} dataset (5,488 Reddit posts), with replacement,
such that the gold anxiety arc had portions with various random amplitudes and slopes (spikiness).\footnote{Sampling without replacement led to the same results, but sampling with replacement allowed reuse of the limited labeled data to produce longer waves. The random arc sampling algorithm is described in the Appendix.}
(Note that sampling is a way to create a new ordering of the posts.)
We continued sampling until we created text streams that were 10,000 sentences long.
We then repeated the procedure 1000 times to create 1000 different text streams (and corresponding gold arcs).
These text streams can now be used to evaluate automatic methods of determining anxiety arcs.
% We repeated the above procedure for bin sizes of 1, 10, 50, 100, 200, 500, and 1000, so that we have data for a wide variety of window sizes.\footnote{}
% In all this resulted in 7,000 text streams and corresponding gold arcs (1000 for each bin size).
% For a given stream of sentences labeled with 0 or 1 (manually provided) anxiety  labels and 
% % of say 50 instances per bin, we compute the gold anxiety score by 
% taking the average of the human-labeled anxiety scores of the instances (sentences or posts) in each of the bins. We then move the window forward by one sentence, compute the average anxiety score
% % of the % human-labeled 
% % emotion scores of the instances 
% in that window (sentnces 2 to k+1) --- which becomes the second point in the gold arc, and so on. Finally, when the window reaches the last sentence in teh stream, the full gold arc is complete.
%% The gold arc is then standardized as before.
%%We will refer to this new text stream generated from a dataset as \textit{[dataset\_name]-dynamic},
%% and the gold arcs created with this process as \textit{dynamic gold arcs}.
The yellow line in Figure \ref{fig:anxarc-avad} (a) shows the beginning portion of one of the gold anxiety arcs (the 1000th arc)
for bin size of 50.

\subsection{Results}

The green line in Figure \ref{fig:anxarc-avad} (a) shows the beginning portion of the predicted anxiety arc when using the lexicon-only method, word associations from WorryWords, and a rolling window with bin size 50.\footnote{Note that bin size is a user specified parameter. Bin size of 50 posts is relatively small and often only used when data is sparse, e.g., tracking character arcs in novels. Results with bigger bin sizes are often better (as more data is used to infer change of emotion across bins) -- as shown ahead.}
Observe that the green line traces the gold 
%dotted 
line quite closely. There are also occasions where it overshoots or fails to reach the gold arc.
The column on the right in the figure lists the average Spearman correlation between the predicted and gold arcs for all 1000 text streams (for the full length of the text streams: 10,000 posts) .
We note that the arc generated using the anxiety lexicon has a  high correlation with the gold arc (greater than 0.9) -- sufficient for most applications. 

Figure \ref{fig:anxarc-avad} (b), (c), and (d) show the same gold anxiety arc but show predicted arcs using valence, arousal, and dominance scores, respectively (taken from the NRC VAD Lexicon). These figures act as control arcs (for comparison with the WorryWords arc). %for the main results obtained with the anxiety lexicon. 
% Observe that the arc generated using the valence lexicon rises when the gold arc falls and falls when the gold arc rises, indicating a negative correlation. The arc for arousal is somewhat close to the gold arc, but not as close as the arc generated using the anxiety lexicon. The arc generated using dominance scores is markedly different from the gold arc.
The column on the right shows the average correlations for the 1000 pairs of gold and predicted arcs.
We see that the gold arc is highly negatively correlated with valence (-0.85). %; or in other words highly correlated with negativeness (the inverse of valence).
 Just in terms of the magnitude of correlation, arousal is closer to anxiety than dominance with anxiety.

  \begin{table}[t]
\small
\vspace*{2mm}
 \centering
%\resizebox{0.28\textwidth}{!}{
    \begin{tabular}{rrr}
    \hline
  \textbf{Window Size}    &\textbf{Spearman} &\textbf{RMSE}\\
     \hline
	    1       &0.554 &1.011\\
        10      &0.767 &0.517\\
	    50      &0.903 &0.204\\
	    100     &0.932 &0.136\\
	    200     &0.946 &0.099\\
        500     &0.955 &0.075\\
        1000    &0.955 &0.073\\
        \hline
    \end{tabular}
 %   }
    \caption{Avg.\@ Spearman corrn.\@ and RMSE of 1K pairs of predicted and gold anxiety arcs for different bin sizes.}
    \label{tab:emooarc-binsizes}
    \vspace*{-4mm}
\end{table}

\noindent {\bf Categorical Emotions:} Figure \ref{fig:anxarc-8emo} in the Appendix shows similar gold and predicted arcs (and overall average correlations) for the gold anxiety arc and predicted arcs generated using 
lexicons for eight other categorical emotions. %: anger, anticipation, disgust, fear, joy, sadness, surprise, and trust. 
The magnitudes of correlation obtained using lexicons for these emotions is lower than that obtained using WorryWords.
% with inverse of joy coming closest, followed by sadness and anticipation.
% the anxiety lexicon (repeated for ease of reference) and 

\noindent {\bf Window Size Effect:} We repeated the experiments described above with many window sizes and found similar overall trends. However, using larger window sizes generally led to higher correlations. Table \ref{tab:emooarc-binsizes} shows these results. In addition to Spearman rank correlation the table also shows another metric of arc--arc closeness: root mean square error (RMSE). The closer the two arcs, the smaller the RMSE. Observe that there is a plateauing effect in closeness after a window size of a few hundred posts (around $\rho = 0.955$ and RMSE $= 0.075$).

% \section{Word Bigrams Associated with Anxiety}

% \begin{table}[]
% \small
%     \centering
%     \begin{tabular}{lrrrr}
%                  &\textbf{anxiety} & \textbf{valence} & \textbf{arousal} & \textbf{dominance} \\
%      \hline
%             \textbf{anxiety}	&1.000		&-0.500		&0.242		&-0.238	\\	
% \textbf{valence}	&-0.500		&1.000		&-0.047		&0.570		\\
% \textbf{arousal}	&0.242		&-0.047		&1.000		&0.274		\\
% \textbf{dominance}	&-0.238		&0.570		&0.274		&1.000\\
%         \hline
%     \end{tabular}
%     \caption{Pearson.}
%     \label{tab:corrP}
% \end{table}

% 1
% Average across runs:
% 	Avg. Spearman correlation: 0.553877376498066
% 	Avg. Pearson correlation: 0.494564366697081
% 	Mean squared error: 1.01087126660584

% 10
% Average across runs:
% 	Avg. Spearman correlation: 0.767193621236274
% 	Avg. Pearson correlation: 0.741537619487827
% 	Mean squared error: 0.516924761024344

% 100
% Average across runs:
% 	Avg. Spearman correlation: 0.931662604540619
% 	Avg. Pearson correlation: 0.931781242758772
% 	Mean squared error: 0.136437514482454

% 200
%   Average across runs:
% 	Avg. Spearman correlation: 0.945650702584876
% 	Avg. Pearson correlation: 0.950457280053169
% 	Mean squared error: 0.099085439893661

% 500
% Average across runs:
% 	Avg. Spearman correlation: 0.954642155758078
% 	Avg. Pearson correlation: 0.962334140349842
% 	Mean squared error: 0.075331719300317

% 1000
% Average across runs:
% 	Avg. Spearman correlation: 0.954084173942215
% 	Avg. Pearson correlation: 0.963771745254524
% 	Mean squared error: 0.0724565094909535

\section{Conclusion}
We created WorryWords --- a large lexicon of word--anxiety association scores with the help of responses to a simple questionnaire, by over 1000 respondents (mainly from the US, but also from India, UK, and Canada).
We showed that the responses led to highly reliable scores. We compared how anxiety association scores correlate with the scores for various other emotions. We determined the rate at which anxiety words are acquired with age; specifically, how from ages 1 to 4 children learn more calmness words than anxiety words, and that trend is reversed age 5 onwards. Finally, we showed how WorryWords can be used to track anxiety in large text streams. We make WorryWords freely available to enable a wide variety of anxiety associated research. 
% (such as in the seven bullets listed in the introduction) in NLP, psychology, social science, etc.

% most notably in understanding and managing anxiety associated disorders, 

\section{Limitations}
\label{sec:limitations}

For limitations associated with the dataset, please refer to the Ethics Statement (\S \ref{sec:ethics}).

The lexicographic experiments in Section 6 make use of existing emotion and Age of Acquisition datasets. The labels for these datasets reflect their annotators, who were mainly US and European English speakers. Thus further experiments with datasets pertaining to other locations and demographics are needed to determine how widely applicable or universal some of these conclusions are. The conclusions form our work should largely be seen as applicable to the demographics of its annotators (and not beyond). 

The emotion arc experiments were done on an existing stress-labeled dataset of Reddit posts. However, the datasets does not correspond to a natural textstream such as tweets about a new government policy or utterances of a character in a novel. (No such dataset has human annotations for anxiety.) However, by creating text streams corresponding to a thousand different emotion arcs from an existing  Reddit dataset, we created a robust anxiety arc gold dataset.

\section{Ethics Statement}
\label{sec:ethics}

Emotions are complex and nuanced mental states. Additionally, each individual expresses emotions, including anxiety, differently through language, which results in large amounts of variation. 
See \citet{Mohammad23ethicslex} for a discussion of good practises and ethical considerations when using emotion lexicons. See \citet{Mohammad22AER} for a broader discussion of ethical considerations relevant to automatic emotion recognition.
We discuss below notable ethical considerations when computationally analyzing anxiety. 

Importantly, WorryWords should not be used as a standalone tool for detecting anxiety disorders. We hope future research, led by clinicians and medical practitioners, will assess the usefulness of WorryWords in mental health applications, in combination with various other sources of information. 

The crowd-sourced task was approved by our Institutional Research Ethics Board. 
Our annotation process stored no information about annotator identity and as such there is no privacy risk to them.  The individual words selected did not pose any risks beyond occasionally reading text on the internet. 
The annotators were free to do as many word annotations as they wished. The instructions included a brief description of the purpose of the task (Figure \ref{fig:det-instr}). 

Any lexical dataset of emotion-association norms entails several ethical considerations. 
% BB We list some notable ones below. 
% Many of these were first introduced in \cite{mohammad2020practical,mohammad2023LexEthics}. We adapted them to anxiety association and added to the discussion.
\vspace*{-1mm}
\begin{itemize}
    \item \textit{Coverage:} We sampled a large number of English words from many sources (which themselves sample from many sources). Yet, it is possible that the words included do not cover all domains, genres, and people of different locations, socio-economic strata, etc. equally. It likely includes more of the vocabulary common in the United States with a socio-economic and educational backgrounds that allow for technology access.
    \vspace*{-1mm}
    \item \textit{Word Senses and Dominant Sense Priors:} Words when used in different senses and contexts may be associated with different degrees of anxiety. The entries in WorryWords are indicative of the anxiety associated with the predominant senses of the words. This is usually not too problematic because most words have a highly dominant main sense (which occurs much more frequently than the other senses). 
% Further, the lexicons only provide information about which emotion association is most likely. 
% Words can be used in contexts where they convey a different emotion.
In specialized domains, some terms might have a different dominant sense than in general usage. Entries in the lexicon for such terms should be appropriately updated or removed. 
\vspace*{-1mm}
    \item \textit{Not Immutable:} The anxiety scores do not indicate an inherent unchangeable attribute. The associations can change with time (e.g., the increase in anxiety associated with \textit{coughing} during a world-wide pandemic), but the dataset entries are largely fixed. They pertain to the time they are created.
    \vspace*{-1mm}
    \item \textit{Socio-Cultural Biases:} The annotations for anxiety capture various human biases. These biases may be systematically different for different socio-cultural groups. Our data was annotated by mostly US and Indian English speakers, but even within the US and India there are many diverse socio-cultural groups.
    Notably, crowd annotators on Amazon Mechanical Turk do not reflect populations at large. In the US for example, they tend to skew towards male, white, and younger people. However, compared to studies that involve just a handful of annotators, crowd annotations benefit from drawing on hundreds and thousands of annotators (such as this work). 
    \vspace*{-1mm}
    \item \textit{Inappropriate Biases:} Our biases impact how we view the world, and some of the biases of an individual may be inappropriate. For example, one may have race or gender-related biases that may percolate subtly into one's notions of anxiety associated with words. 
    Our dataset curation was careful to avoid words from problematic sources. We also ask people annotate terms based on what most English speakers think (as opposed to what they themselves think). This helps to some extent, but it can still capture historical anxiety associations with certain identity groups. This can  be useful for some socio-cultural studies; but we also caution that anxiety association with identity groups be carefully contextualized to avoid false conclusions.    
    \vspace*{-1mm}
    \item \textit{Perceptions (not “right” or “correct” labels):} Our goal here was to identify common perceptions of anxiety association. These are not meant to be ``correct'' or ``right'' answers, but rather what the majority of the annotators believe based on their intuitions of the English language.
    \vspace*{-1mm}
    \item It is more appropriate to make claims about anxiety word usage rather than anxiety of the speakers. For example, {\it `the use of anxiety words grew by 20\%'} rather than {\it `anxiety grew by 20\%'}. 
A marked increase in anxiety words is likely an indication that anxiety increased, but there is no evidence that anxiety increased by 20\%.\\[-16pt]
\item Comparative analyses can be much more useful than stand-alone analyses. Often, anxiety word counts on their own are not useful. 
For example, {\it `the use of anxiety words grew by 20\% when compared to [data from last year, data from a different person, etc.]'} is more useful than saying {\it `on average, 5 anxiety words were used in every 100 words'}.\\[-16pt]
\item Inferences drawn from larger amounts of text are often more reliable than those drawn from small amounts of text.
 For example, {\it `the use of anxiety words grew by 20\%'} is informative when determined from hundreds, thousands, tens of thousands, or more instances. Do not draw inferences about a single sentence or utterance from the anxiety associations of its constituent words.     
\end{itemize}
\noindent We recommend careful reflection of ethical considerations relevant for the specific context of deployment when using WorryWords.

\section*{Acknowledgments}
Many thanks to Sophie Wu and Tara Small for helpful discussions.

\bibliography{anthology,custom}
\bibliographystyle{acl_natbib}

% \newpage
\appendix

\section{APPENDIX}
\label{appendix-a}

\subsection{Sources of Terms Selected for Annotation}

WorryWords includes terms selected from these resources:
\begin{itemize}
    \item All terms in the NRC Emotion Lexicon v1 \cite{MohammadT13,MohammadT10}: It includes about 14,000 common English words with labels indicating whether they are associated with any of the eight notable discrete emotions identified by \newcite{Plutchik80}: 
anger, anticipation, disgust, fear, joy, sadness, surprise, and trust.
%The NRC lexicon terms were in turn chosen by taking the content words that occur frequently in the Google n-gram corpus \cite{BrantsF06}.
    \item All terms in the NRC VAD Lexicon v1 \cite{mohammad-2018-obtaining}: It includes about 20,000 common English terms. Notably, it also includes terms common in X (formerly Twitter). Posts on X include non-standard language such as creatively spelled words ({\it happee}), hashtags ({\it \#takingastand, \#lonely}) and conjoined words ({\it loveumom}).  
% About 1000 high-frequency content terms, including emoticons, 
These were taken from the Twitter Emotion Corpus \cite{COIN:COIN12024}.

    \item Terms in the Prevalence dataset \cite{brysbaert2019word}  
    that are marked as known to 70\% or more respondents:
    % Prevalence score greater than 0.5: 
    This dataset has prevalence scores (how widely a word is known by English speakers), determined directly by asking people, for 62,000 lemmas. We included a term if it was marked as known to at least 70\% of the people who provided responses for the term. 
    % There were 45,222 such terms in total.
    From this set we removed terms that are common person names or city names.
    
    % (taken from publicly available lists). 
    
    % \footnote{Percentage scores of number of people who know a word are probit transformed (Brysbaert et al., 2016b) to create the final prevalence scores. The probit function translates percentages known to z values on the basis of the cumulative normal distribution. That is, a word known by 2.5\% of the participants would have a word prevalence of $-1.96$; a word known by 97.5\% of the participants would have a prevalence of $+1.96$.} We included all terms with prevalence scores $> 0.5$ that were not common person or city names. There were blah such terms.

    % Because the distribution of percentages known was very right-skewed and did not differentiate much between well-known words, it was useful to apply a probit transformation to the percentages (Brysbaert et al., 2016b). The probit function translates percentages known to z values on the basis of the cumulative normal distribution. That is, a word known by 2.5% of the participants would have a word prevalence of – 1.96; a word known by 97.5% of the participants would have a prevalence of + 1.96. Because a word known by 0% of participants would return a prevalence score of – ∞ and a percentage known of 100% would return a prevalence score of + ∞, the range was reduced to percentages known from 0.5% (prevalence = – 2.576) to 99.5% (prevalence = + 2.576).
\end{itemize}

\begin{table}[t]
\centering
% \begin{center}
{\small
\begin{tabular}{ll}
\hline
\bf Word	&\bf Anxiety Score \\\hline
suffocative	&3\\
smolder	&3\\
exploder	&3\\
exterminate	&2.71\\
ghastly	&2.66\\
% frightfulness	&2.66\\
haunted	&2.62\\
castrate	&2.62\\
% embalmment	&2.55\\
abusive	&2.5\\
manic	&2.4\\
knifelike	&2.4\\
coercion	&2.33\\
% hallucinating	&2.33\\
conniving	&2.16\\
affliction	&2.14\\
frustrate	&2.05\\
unforeseeable	&2\\
grueling	&2\\
snarling	&1.88\\
obtrusion	&1.88\\
hysterectomy	&1.87\\
scum	&1.85\\
reinjure	&1.83\\
riskily	&1.72\\
downheartedly	&1.71\\
jailhouse	&1.71\\
frostbitten	&1.66\\
stifled	&1.62\\
disturbed	&1.57\\
projectile	&1.55\\
disquieted	&1.55\\
centaur	&1.33\\
brusquely	&1.25\\
psychosomatic	&1.12\\
escapist	&1\\
discordance	&0.8\\
propagandize	&0.75\\
brandy	&0.14\\
ceramics	&0.12\\
photosphere	&0\\
rhinoceros	&0\\
paraphrase	&0\\
forkless	&0\\
witting	&0\\
gemini	&-0.1\\
swordswoman	&-0.13\\
canto	&-0.41\\
roller	&-0.5\\
motorcycle	&-0.71\\
semiprofessional	&-0.8\\
socialization	&-0.8\\
cleanness	&-1.55\\
coldblooded	&-1.57\\
fantail	&-1.6\\
conformed	&-1.71\\
unscratched	&-1.75\\
journalistic	&-1.8\\
fireless	&-1.85\\
reaffirmation	&-2\\
aristocratic	&-2.1\\
limberness	&-2.5\\
unblemished	&-2.62\\
lullaby &-2.79\\
delightfully    &-3\\
generous    &-3\\
\hline
\end{tabular}
\caption{\label{tab:examples} {Randomly sampled terms and their anxiety-association score from WorryWords.}}
}
% \vspace*{-3mm}
% \end{center}
\end{table}

\subsection{AMT Questionnaire}
\begin{figure*}[t]
	     \centering
	     \includegraphics[width=0.85\textwidth]{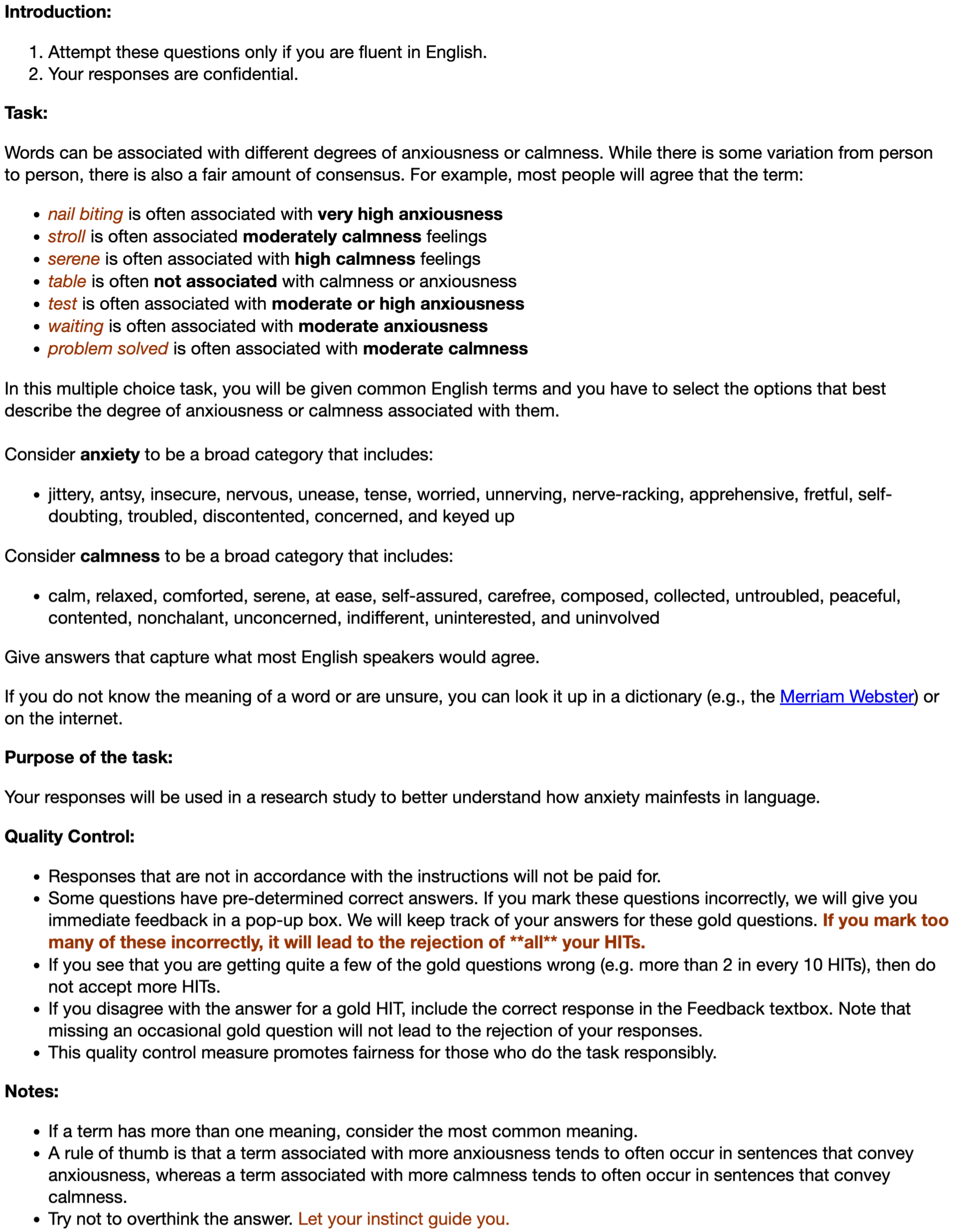}
	     \caption{Questionnaire: Detailed instructions.}
	     \label{fig:det-instr}

	 \end{figure*}

\begin{figure*}[t]
	     \centering
	     \includegraphics[width=0.65\textwidth]{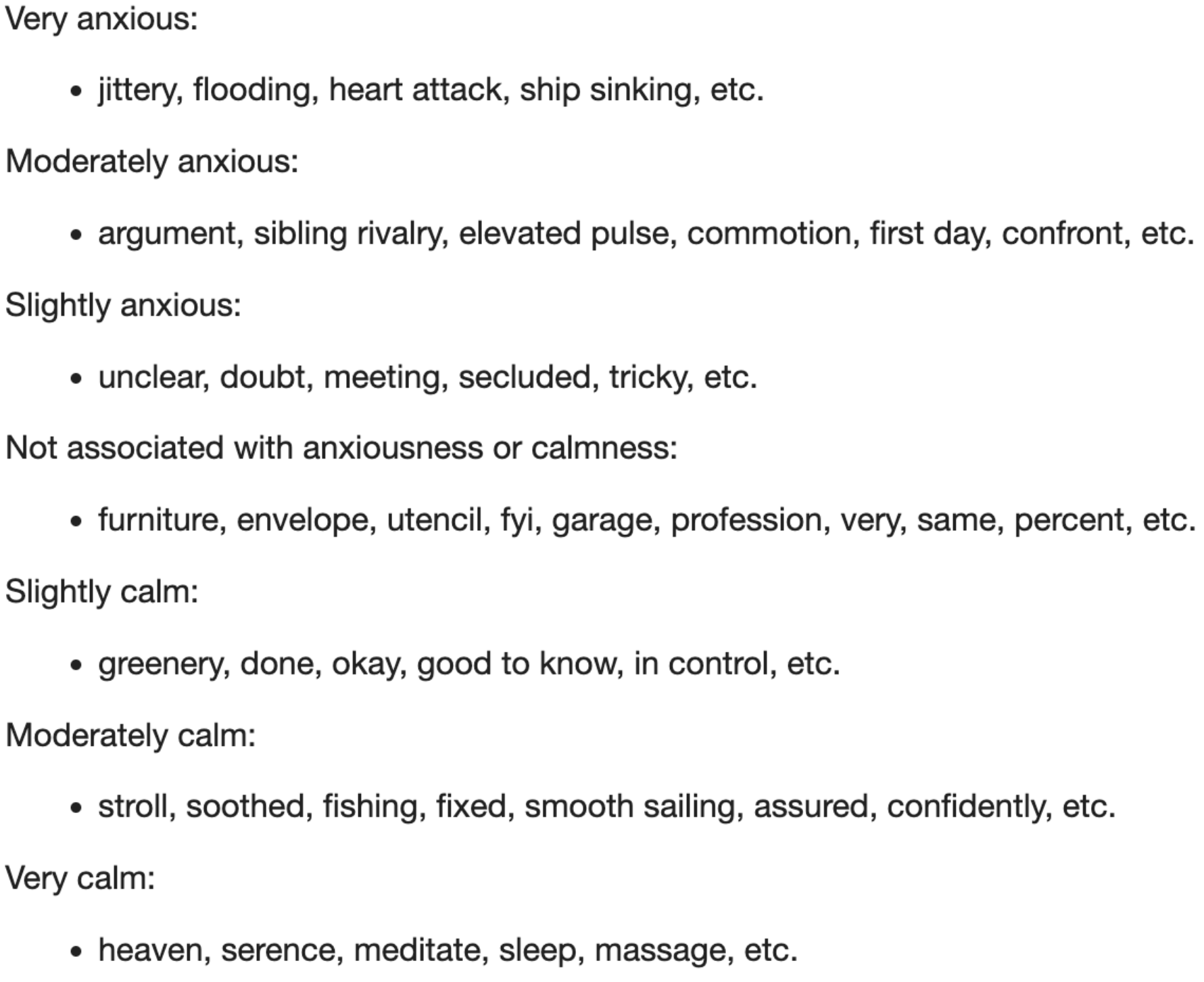}
	     \caption{Questionnaire: Examples.}
	     \label{fig:q-examples}
	 \end{figure*}

\begin{figure*}[t]
	     \centering
	     \includegraphics[width=0.6\textwidth]{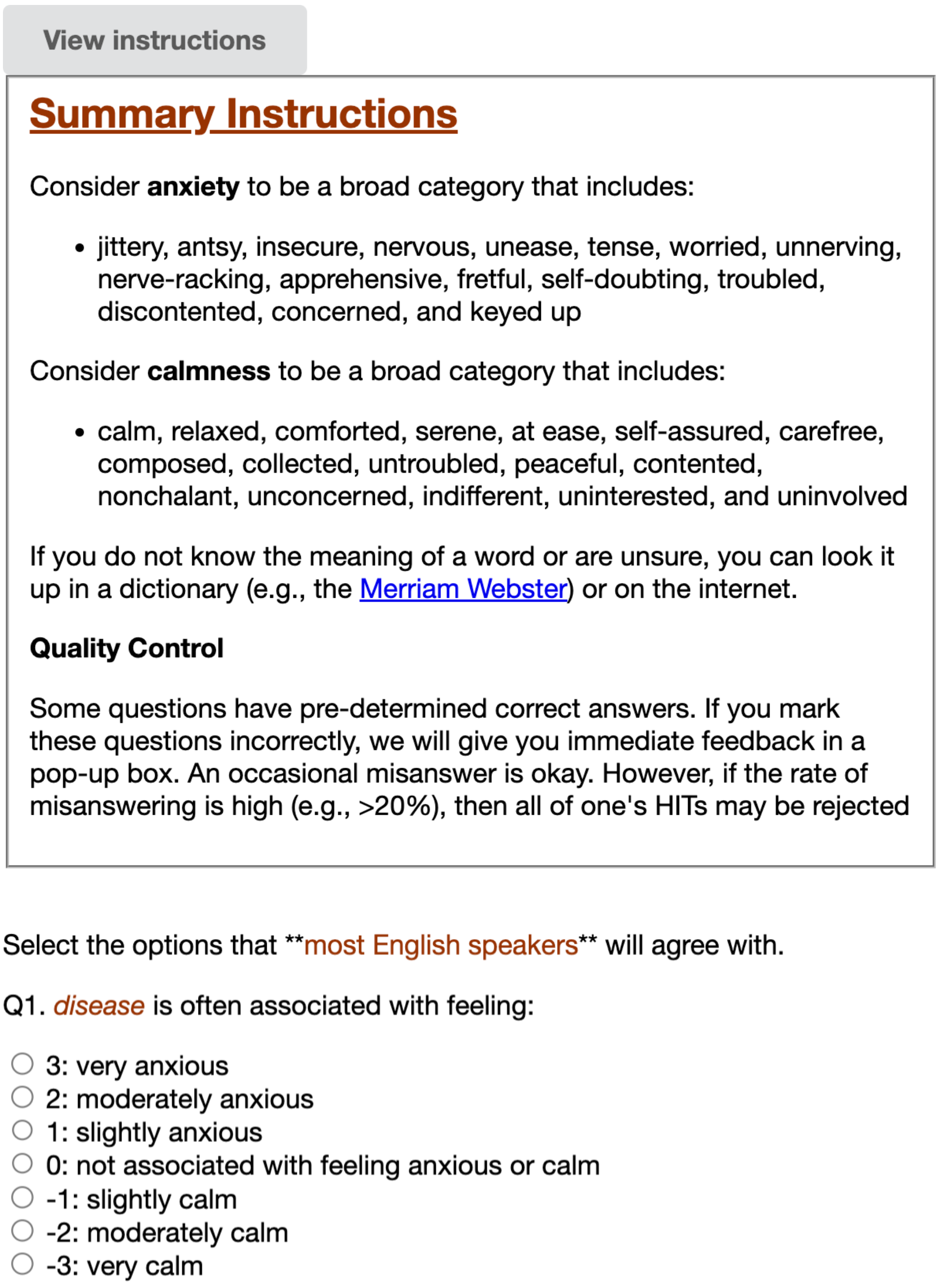}
	     \caption{Questionnaire: Sample Question.}
	     \label{fig:q-main}
	 \end{figure*}

Screenshots of the detailed instructions, examples, and sample question presented to the annotators are shown in Figures \ref{fig:det-instr}, \ref{fig:q-examples}, and \ref{fig:q-main}, respectively.

\subsection{Distribution of WorryWords}

WorryWords is made freely available on the project website as a compressed file. Terms of use will require that users not re-distribute the file and not post any form of the lexicon on the web. This is to prevent the resource being included in the data scrape fed to a large language model. 
See full list of terms of use at the project home page.
Table \ref{tab:examples} shows entries for a random sample of words from WorryWords.  

\clearpage

\newpage

%  \item SUBTLEX \cite{new2007use}:

% \item All 4,206 terms in the positive and negative lists of the General Inquirer \cite{Stone66}.\\[-19pt]
% \item All 1,061 terms listed in ANEW \cite{bradley1999affective}.\\[-19pt]
% \item All 13,915 terms listed in the \newcite{warriner2013norms} lexicon.\\[-19pt]
% \item 520 words from the Roget's Thesaurus categories corresponding to the eight basic Plutchik emotions.\footnote{http://www.gutenberg.org/ebooks/10681}\\[-19pt]
% Note that this set of terms includes both terms that are more common in social media communication (for example, {\it :), soannoyed, grrrrr, stfu}, and {\it thx})
% as well as regular English words.\footnote{Some of the terms included from the Twitter source were deliberate spelling variations of English words, for example, {\it bluddy} and {\it sux}.}\\[-22pt]

%  \begin{table}[t!]
% \begin{center}
% {\small
% \begin{tabular}{lclc}
% \hline
% \bf Word	&\bf Score$\uparrow$	&\bf Word	&\bf Score$\downarrow$ \\\hline
% \textit{infuriated}  & 3                     & \textit{applaud}         & -3\\
% \textit{traumatized} & 3                     & \textit{benedictional}   & -3\\
% \textit{warcrimes}   & 3                     & \textit{home}            & -3\\
% \textit{homicidal}   & 3                     & \textit{harmlessness}    & -3\\
% \textit{panicking}   & 3                     & \textit{melodiously}     & -3\\
% \hline
% \end{tabular}
% \caption{\label{tab:examples} {Example terms with the highest ($\uparrow$) and lowest ($\downarrow$) anxiety scores in WorryWords.}}
% }
% % \vspace*{-3mm}
% \end{center}
% \end{table}

  \clearpage

 \begin{figure*}[t]
	     \centering
	     \includegraphics[width=0.85\textwidth]{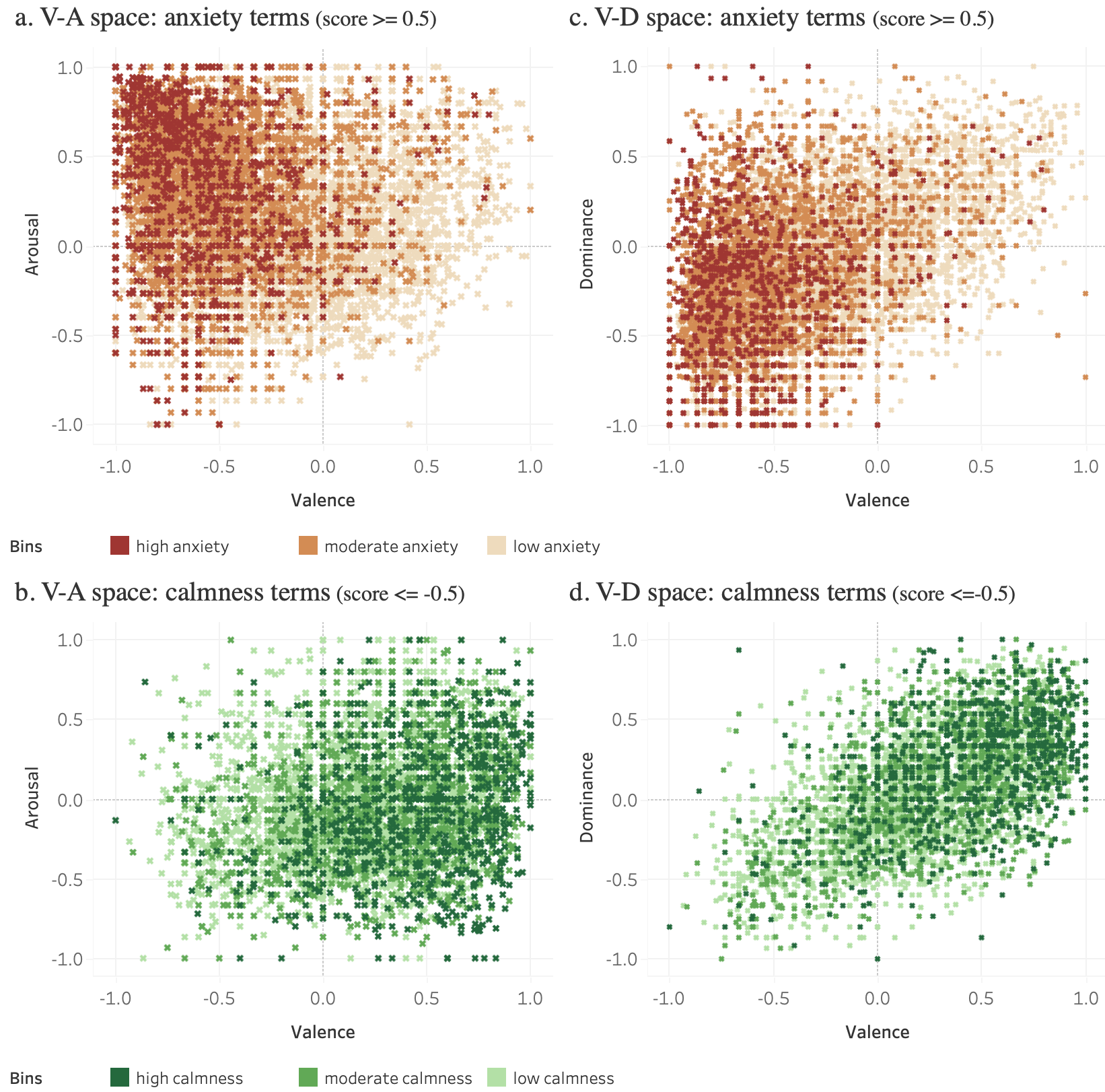}
	     \caption{WorryWords terms in the Valence--Arousal (a and b) and Valence--Dominance (c and d) spaces.}
	     \label{fig:anx-scat-vad}

	 \end{figure*}

 \begin{figure*}[t]
	     \centering
	     \includegraphics[width=\textwidth]{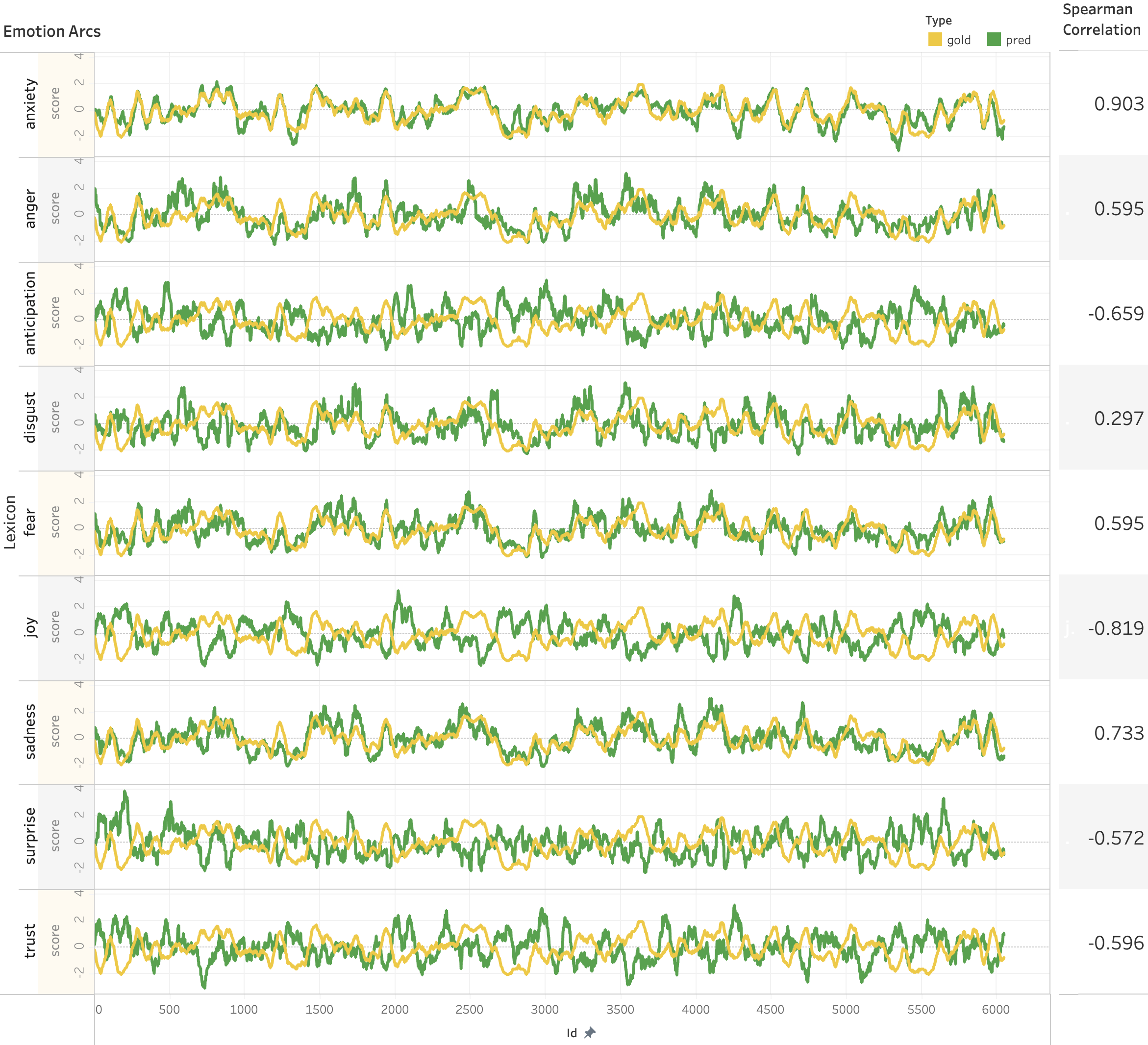}
	     \caption{Predicted (green) anxiety arcs when applying nine different types of association scores to one of the test data text streams: 
1. associations with anxiety (from WorryWords); 2--9. associations with each of the eight Plutchik emotions (from NRC Emotion Intensity Lexicon). The gold (yellow) arc pertaining to the same text stream is repeated in all nine figures for reference. Bin size used to create the arcs: 50 posts. On the right is the average correlation between 1000 pairs of gold and predicted arcs for each case.}
	     \label{fig:anxarc-8emo}

	 \end{figure*}

% \begin{figure*}
% \centering
% \includegraphics[width=\textwidth]{latex/figures/reddit-anx-vad-arcs.png}
% \caption{A .}
% \vspace*{-5mm}
% \label{fig:ued-example}
% \end{figure*}

\begin{table*}[t]
\small
    \centering
    \begin{tabular}{l|rrrrrrrrr}
              \hline
	&\textbf{anxiety} &\textbf{anger}	&\textbf{anticipn.}		&\textbf{disgust}	&\textbf{fear}	&\textbf{joy}	&\textbf{sadness}	&\textbf{surprise}	&\textbf{trust}\\
     \hline
\textbf{anxiety}	&1.000		&0.336		&-0.011		&0.279		&0.357		&-0.132		&0.272		&0.082		&-0.137		\\
\textbf{anger}	&0.336		&1.000		&0.025		&0.432		&0.461		&-0.023		&0.372		&0.117		&-0.048		\\
\textbf{anticipn.}	&-0.011		&0.025		&1.000		&-0.014		&0.074		&0.432		&0.020		&0.276		&0.252		\\
\textbf{disgust}	&0.279		&0.432		&-0.014		&1.000		&0.328		&-0.036		&0.323		&0.059		&-0.054		\\
\textbf{fear}	&0.357		&0.461		&0.074		&0.328		&1.000		&-0.017		&0.439		&0.167		&-0.035		\\
\textbf{joy}	&-0.132		&-0.023		&0.432		&-0.036		&-0.017		&1.000		&-0.014		&0.283		&0.386		\\
\textbf{sadness}	&0.272		&0.372		&0.020		&0.323		&0.439		&-0.014		&1.000		&0.088		&-0.047		\\
\textbf{surprise}	&0.082		&0.117		&0.276		&0.059		&0.167		&0.283		&0.088		&1.000		&0.106		\\
\textbf{trust}	&-0.137		&-0.048		&0.252		&-0.054		&-0.035		&0.386		&-0.047		&0.106		&1.000		\\
        \hline
    \end{tabular}
    \caption{Spearman rank correlation between the word--emotion association scores for various emotion pairs. For a given anxiety--emotion pair, the correlation is computed for terms that are included in corresponding emotion lexicons.}
    \label{tab:corrP-anx-8emo}
\end{table*}

\newpage

\subsection{Random Arc Sampling Algorithm}

In order to create a text stream (described in Section 6) such that the gold arc has various components of random ups and downs of varying slopes, we sample posts from a dataset of anxiety (1) and no-anxiety (0) posts, with replacement, in short bursts as follows.
For each burst (sub-arc), we first randomly determine a slope and number of posts to make up the sub-arc. Then we sample posts from the source datasets accordingly. For example, sampling only anxiety posts will lead to a slope of 90 degrees, sampling only no-anxiety posts will lead to a slope of -90 degrees, and sampling a mixture will lead to a slope between -90 and +90 (depending on the proportion of anxiety and no-anxiety posts). We continue sampling until the randomly chosen length of the subarc is reached. Then the whole procedure is repeated with new randomly generated slopes and lengths of subarcs until the text stream includes 10,000 posts.  

\subsection{Supplementary Figures and Tables}

Figure \ref{fig:anx-scat-vad} shows plots of the WorryWords terms on valence--arousal and valence--dominance spaces (described in Section 6.1).
Table \ref{tab:corrP-anx-8emo} shows the Spearman rank correlation between the word--emotion association scores for various emotion pairs (discussed in Section 6.1). 
%% For a given emotion pair, the correlation is computed for terms that are included in corresponding emotion lexicons.

Figure \ref{fig:anxarc-8emo} shows predicted (green) anxiety arcs when applying nine different emotion association scores to one of the test data text streams (described in Section 7).

\subsection{Computational Resources and Carbon Footprint}
A nice advantage of using simple lexicon-based approaches is the low carbon footprint and computational resources required. All of the experiments described in the paper were conducted on a regular personal laptop.

\end{document}